\documentclass[10pt,twocolumn,letterpaper]{article}

\usepackage[pagenumbers]{cvpr} 

\usepackage{graphicx}
\usepackage{amsmath}
\usepackage{amssymb}
\usepackage{booktabs}
\usepackage{multirow}

\usepackage[pagebackref,breaklinks,colorlinks]{hyperref}

\usepackage[capitalize]{cleveref}
\crefname{section}{Sec.}{Secs.}
\Crefname{section}{Section}{Sections}
\Crefname{table}{Table}{Tables}
\crefname{table}{Tab.}{Tabs.}

\def\cvprPaperID{7954} 
\def\confName{CVPR}
\def\confYear{2023}

\begin{document}


\title{FashionSAP: Symbols and Attributes Prompt for Fine-grained Fashion Vision-Language Pre-training}
\author{Yunpeng Han\textsuperscript{1},Lisai Zhang\textsuperscript{1},Qingcai Chen\textsuperscript{1,2}\thanks{Corresponding author}, Zhijian Chen\textsuperscript{3},Zhonghua Li\textsuperscript{3},Jianxin Yang\textsuperscript{3},Zhao Cao\textsuperscript{3}\\
\textsuperscript{1}Harbin Institute of Technology Shenzhen, China\\
\textsuperscript{2}PengCheng Laboratory, Shenzhen China, 
\textsuperscript{3}Huawei Technologies Co., Ltd\\
{\tt\small hanyunpeng.hyp@gmail.com, lisaizhang@foxmail.com, qingcai.chen@hit.edu.cn}\\
{\tt \small \{chenzhijian13, lizhonghua3, yangjianxin4, caozhao1\}@huawei.com}
}
\maketitle

\begin{abstract}
   Fashion vision-language pre-training models have shown efficacy for a wide range of downstream tasks. However, general vision-language pre-training models pay less attention to fine-grained domain features, while these features are important in distinguishing the specific domain tasks from general tasks. We propose a method for fine-grained fashion vision-language pre-training based on fashion \textbf{S}ymbols and \textbf{A}ttributes \textbf{P}rompt (FashionSAP) to model fine-grained multi-modalities fashion attributes and characteristics. Firstly, we propose the fashion symbols, a novel abstract fashion concept layer, to represent different fashion items and to generalize various kinds of fine-grained fashion features, making modelling fine-grained attributes more effective. Secondly, the attributes prompt method is proposed to make the model learn specific attributes of fashion items explicitly. We design proper prompt templates according to the format of fashion data. Comprehensive experiments are conducted on two public fashion benchmarks, i.e., FashionGen and FashionIQ, and FashionSAP gets SOTA performances for four popular fashion tasks. The ablation study also shows the proposed abstract fashion symbols, and the attribute prompt method enables the model to acquire fine-grained semantics in the fashion domain effectively. The obvious performance gains from FashionSAP provide a new baseline for future fashion task research.\footnote{The source code is available at \url{https://github.com/hssip/FashionSAP}}
\end{abstract}

\section{Introduction}
\label{sec:intro}

\begin{table*}
\linespread{1.15} 
  \centering 
  \setlength\tabcolsep{13pt}
  \begin{tabular}{rll}
    \toprule
    Fashion Symbols & Categories & Definition Rules\\
    \midrule
    \textit{TOPS} & tops, shirt, polo, sweater, ... & upper body \\
    \textit{DRESSES} & dress, suit, shift, ...& up-to-lower body \\
    \textit{SKIRTS} & skirt, sarong, slit, kilt, ... & lower body\\
    \textit{COATS} & jacket, parka, blazer, duffle, ... & associated with others\\
    \textit{PANTS} & jeans, shorts, breeches, ...  & lower body\\
    \textit{SHOES} & boots, sneakers, pump, loafers, ... & feet\\
    \textit{BAGS} & clutches, pouches, wristlet, ... & bag \& decorative\\
    \textit{ACCESSORIES} & \begin{tabular}[l]{@{}l@{}l}
    ring, sunglasses, accessories, \\ hat, necklace, .... \end{tabular} & decorative \& optional\\
    \textit{OTHERS} & swim-wear, lingerie, lounge-wear, ..., & -\\
    \bottomrule
  \end{tabular}
  \caption{Fashion symbols and corresponding categories with definition rules.}
  \label{tab:fashionsymbol}
\end{table*}

Vision-Language pre-training (VLP) attracts wide attention\cite{ALBEF,BLIP,yao2021filip,li2020oscar, soho} as a foundation for general multi-modal tasks. 
VLP methods aim at learning multimodal knowledge from large-scale text and image pairs data containing common objects in daily life. For example, MSCOCO\cite{MSCOCO2014}, a public vision-language benchmark, is introduced with common object labels.
The fashion domain is an important application of VLP methods, where the online retail market needs retrieval and recommendation services. To satisfy such requirements, the VLP model needs to learn high-quality representations containing fine-grained attributes from the fashion data.
    %

\begin{figure}
  \centering \small
  \includegraphics[width=0.9\linewidth]{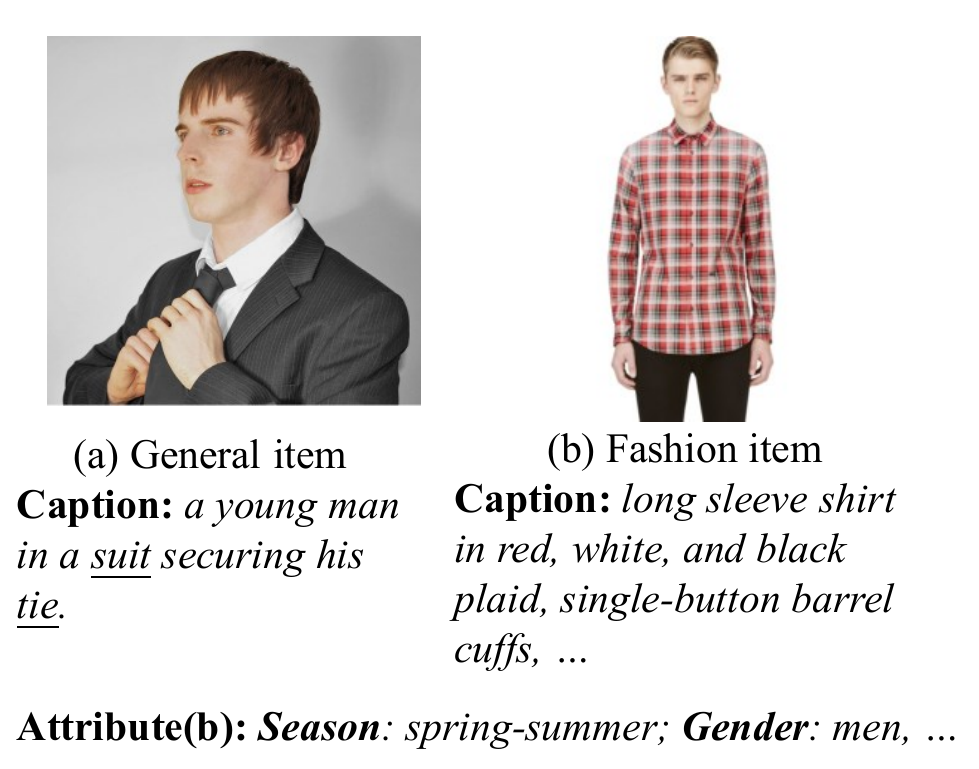}
  \caption{Two text-image instances from general (a) and fashion domain (b). The captions of the general domain only describe object-level (underlined words) image content, while fashion domain captions emphasise attribute-level semantics.}
  \label{fig:twoinstances}
\end{figure}

Many works have adapted general VLP models to fashion tasks directly. However, the general pre-training models are not effective for learning fashion knowledge to describe fashion items comprehensively, as the fashion descriptions are usually associated with fine-grained attribute-level features. As illustrated in \cref{fig:twoinstances}, the description text of a fashion item (right)  refers to fine-grained attributes like long sleeves, while such features are ignored by the descriptions from general vision-language data (left). 
Moreover, the public fashion producers from fashion platforms attach great importance to some definite attributes (\eg, season, gender) of fashion data and especially provide the fashion attributes annotations. However, these high-quality attributes are highly neglected by existing fashion VLP models. It is important for fashion VLP models to focus on these fine-grained attributes and learn fashion-specific knowledge.

Fashion attributes describe not only item details but also the overall item features. The category for fashion items is an essential attribute highlighted by many benchmark datasets\cite{A100,fashiongen,facad}. We notice that categories have a deep correlation to fine-grained attributes, although they describe the general information of a fashion item. For example, the length is an important attribute for both pants and jeans, while it is rarely mentioned in the description of a pair of shoes. However, most existing fashion VLP methods neglect the importance of the relationship between similar categories. In this paper, we explore the usage of category attributes as a global concept layer during pre-training. According to the human description of a fashion product, we believe categories declare the basic understanding of a fashion product. Therefore, we attach the fashion category to the beginning of captions to guide the representation learning. Since the fashion products are designed for the decoration of people, we summarize nine fashion symbols corresponding to human body parts, as shown in \cref{tab:fashionsymbol}, to unify all the categories of fashion items.  

We propose a method for the fashion domain to learn fine-grained semantics. This method is able to capture the similarity of fine-grained features based on fashion symbols and learn explicit fine-grained fashion attributes by the prompt. Our method gets the SOTA performance for four popular fashion tasks on the two public datasets, and the obvious performance gains provide new baselines for further research. 
Our main contributions are summarized below:
\begin{itemize}
    \item An effective fine-grained vision-language pre-training model is proposed to learn the attribute-level fashion knowledge. 
    \item An abstract fashion concept layer is proposed, and 9 fashion symbols are summarized to represent various fashion concepts according to their similarities on body parts and product functions. 
    \item The attributes prompt method enables the cross-modalities pre-training model to explicitly learn fine-grained fashion characteristics. 
\end{itemize}

\section{Related Work}
\label{sec:relawork}

\textbf{Vision-Language Pre-training}
The pre-training of the vision-language model has been used in many works\cite{ALBEF,BLIP,li2020oscar,zhang2021vinvl, soho}. The structure of the VLP model mainly includes two types, single-stream and two-stream. The single-stream models \cite{zhang2021vinvl,li2020oscar} generate the image and text into preliminary representations and concatenate them so that they can interact with each other in a unified model (\eg transformer \cite{transformer}). Two-stream models \cite{CLIP,ALIGN, hashcode} try to encode text and image respectively and the features interact with each other through semantic alignment tasks. Some works\cite{ALBEF,BLIP,yao2021filip, vldeformer} combine single-stream and two-stream by designing multi-step  semantic alignment tasks. The backbones for text and image encoder refer to the stricture of unimodal\cite{BERT,resnet,vit}. The one-steam models usually perform better than two-stream models, while the latter is better than the former in time complexity. We design a model combined with one-stream and two-stream to adapt to the fashion tasks.

\textbf{Vision-Language Model for Fashion}
Tasks in the fashion domain include the retrieval, match and generation of cross-modal \cite{polyvore} similar to the general vision-language. There are also many datasets collected and released for fashion tasks  \cite{fashiongen,fashioniq,polyvore,polyvoreoutfit,liu2016deepfashion,facad}. KaleidoBERT\cite{fashionbert} designs multiple stages to refine the salient features of fashion items by utilizing multiple single-task frameworks. FashionViL\cite{han2022fashionvil} uses an end-to-end framework to pre-train the model in multiple single-tasks by referring to the general vision-language model. These works try to use attributes of fashion items by attaching all the category attributes to the same classification task. There are also some works aiming at specific fashion tasks \cite{goenka2022fashionvlp,yu2022commercemm,lee2021cosmo,val,composing, EICLIP} by setting a variety of gating and route structures for the latent features of fashion items. 
The exact representation of each attribute respectively is essential for fashion models. We propose a model that can obtain latent features and knowledge in the fashion domain at the pre-training stage.

\textbf{Prompt Learning}
Prompt learning is an effective method to transfer the pre-training model to accomplish downstream tasks in Natural Language Processing\cite{pppp,lidongfang,schick2021exploiting,tam2021improving,sanh2021multitask}. It can utilize the knowledge from the large-scale pre-training model in low-resource scenarios by appropriate prompts. 
The expression of the prompting is the crucial aspect in task transfer\cite{logan2022cutting,lester2021power} as proper trigger words can activate the knowledge from the pre-training model better. Prompted-base methods are also used in task-aimed model training \cite{lidongfang,tam2021improving} for utilizing the resource in the task. These methods diversify a single instance from multiple perspectives into multiple instances. There are also some works applying prompt learning to multi-modal. We design two prompt templates from the text side for the adaptation of different kinds of attributes. \cite{promptmultimodal,yao2021cpt}.

\begin{figure*}
  \centering
  \includegraphics[width=0.9\textwidth]{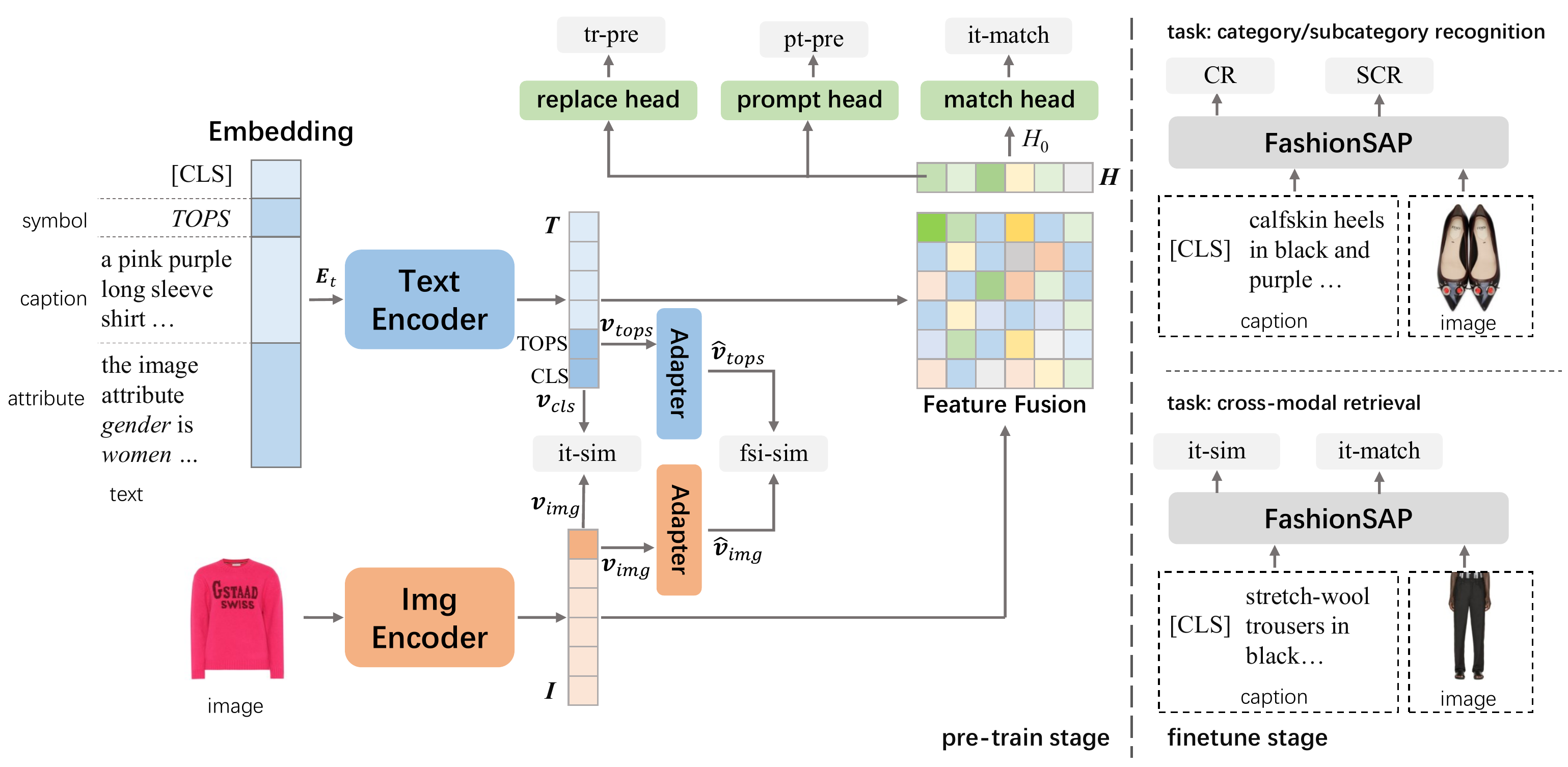}
  \caption{An overview of the FashionSAP framework. The fashion symbol($fsis$ task), attribute prompt ($ptp$ task) and token replace($trp$ task) are all removed in finetune stage.}
  \label{fig:modeloverview}
\end{figure*}
\section{Methodology}
\label{sec:mehtod}

In this section, we first introduce the preliminary of fashion symbols and attributes prompt in \cref{sec:preliminary}. Then we describe the architecture of FashionSAP in \cref{subsec:FSPA} network. Afterward, we elaborate on five pre-training tasks in \cref{subsec:pretraintask}. 

\subsection{Preliminary}
\label{sec:preliminary}
\subsubsection{Fashion Symbols Definition}
\label{subsec:symbols}

The category is an essential attribute of a fashion item. However, the categories terms in different datasets are various. For example, the widely used FashionIQ \cite{fashioniq} provides $3$ kinds of categories while $48$ in FashionGen\cite{fashiongen}. To address the problem, we propose a concept semantic layer to embed similar category terms into the same fashion symbol. The symbols are defined by the following rules:
\begin{enumerate}
    \item \textbf{Body Part}: fashion items that are associated with a specific part of the human body.
    \item \textbf{Function}: fashion items that are optionally used for decoration and can be dressed on multiple body parts.
\end{enumerate}

For the datasets in this paper, we propose nine symbols to summarize different categories of fashion items. As shown in \cref{tab:fashionsymbol}, the fashion symbols \textit{PANTS}, \textit{SKIRTS}, \textit{SHOES}, \textit{BAGS} have their unique features. \textit{TOPS} is a kind of upper clothing that can be worn independently. \textit{DRESSES} can cover the whole body and exist independently. \textit{COATS} represents the outwear usually worn with other clothing. \textit{ACCESSORIES} represents the accessories that aim to enhance the whole outfit but are not necessary for a basic outfit. \textit{OTHERS} includes fashion items that do not appear in everyday dressing and public occasions.  
We use an embedding layer to learn the representation of these fashion symbols as shown in \cref{fig:modeloverview}. We enumerate all categories and corresponding fashion symbols in practice.

\subsubsection{Fine-grained Attribute Prompt}
\label{subsec:PFA}
Existing works suggest that the fashion items are usually annotated from multiple perspectives\cite{A100}. Most benchmark datasets\cite{fashiongen, fashioniq, facad,liu2016deepfashion} focus on fine-grained attributes when annotating fashion items. However, most general vision-language models focus on object-level semantics and seldom pay attention to attribute-level semantics, which contain many fine-grained characteristics for fashion items. Therefore, we propose a method to utilize these fine-grained attributes by prompt.

The attribute format of \textit{key-value} is concordant with the prompt format of \textit{description-value} \cite{schick2021exploiting,tam2021improving}. According to this schema, we encode fine-grained fashion attributes in sequence format so that our model can capture the inner interaction between name and value.  Attributes prompt tells the model precisely the ownership between attribute value and name to utilize the latent semantics from the language model.
We design two prompt templates to tackle the diversity of patterns of fashion attributes. The first template covers the enumerable attributes. 
This kind of attribute has a textual name and the enumerable value from a defined finite set, where each attribute has a unique value. 
The first template is:
\begin{flushleft}
  $\mathcal{T}_e = \verb'the image attribute [An] is [Au]'$
\end{flushleft}
where \verb'[An]' is the slot to be filled with the attribute name, and \verb'[Au]' is filled with the attribute value.
Another template covers the binary attributes. This attribute annotates a fashion item with a one-hot vector with binary to illustrate whether the fashion item has a certain feature, \eg \verb'{red, pure cotton}'. For binary attributes, the template is:
\begin{flushleft}
  $\mathcal{T}_b = \verb'is image attribute [Ab]? [As]'$
\end{flushleft}
where  \verb'[Ab]' is filled with binary attribute label, \verb'[As]' is filled with positive answer word \verb'yes' or negative \verb'no' as attribute value. We concatenate $\mathcal{T}_e$ or $\mathcal{T}_b$ to the tail of the caption tokens during pre-training stage. 

\subsection{Model Architecture}
\label{subsec:FSPA}

As illustrated in \cref{fig:modeloverview}, FashionSAP consists of an image encoder, a text encoder and a feature fusion module. An image is encoded to $\boldsymbol{I}$,
$$\boldsymbol{I}=\{\boldsymbol{v}_{img}, \boldsymbol{v}_{i_0},  \boldsymbol{v}_{i_1}, \boldsymbol{v}_{i_2}, ..., \boldsymbol{v}_{i_N}\}  \in \mathbb{R}^{(i_N+1) \times d}$$
where $ \boldsymbol{v}_{i}$ is a feature vector of a patch of the image generated by IE, $d$ is the dimension of latent semantic space and $i_N$ is the number of patches of the input image. We concatenate the fashion symbol between BERT token \verb'[CLS]' and fashion text to form a new text sequence shown in the upper-left of \cref{fig:modeloverview}. The text sequence is embedded to $\boldsymbol{E}_t$,
 $$\boldsymbol{E}_t = \{\boldsymbol{e}_{cls}, \boldsymbol{e}_{symbol}, \boldsymbol{e}_{t_0}, ..., \boldsymbol{e}_{t_N}\} \in  \mathbb{R}^{(t_N+2) \times d_e}$$
where $t_N$ is the length of fashion text tokens sequence and $d_e$ is the dimension of text embedding space. The embedding $\boldsymbol{E}_t$ is encoded into $\boldsymbol{T}$,
$$\boldsymbol{T}=\{\boldsymbol{v}_{cls},  \boldsymbol{v}_{symbol}, \boldsymbol{v}_{t_0}, ..., \boldsymbol{v}_{t_N}\} \in \mathbb{R}^{(t_N + 2) \times d}$$
For the case in \cref{fig:modeloverview}, the $\boldsymbol{e}_{symbol}$ is specific to $\boldsymbol{e}_{tops}$ and $\boldsymbol{v}_{symbol}$ is specific to $\boldsymbol{v}_{tops}$. 

Then FashionSAP uses a feature fusion module to fuse the features from the text and image into hybrid feature $\boldsymbol{H}$.
The feature fusion module is implemented as multiple cross-attention layers from transformer\cite{transformer}. The feature of \textit{k-th} cross-attention layer is calculated as \cref{equca}
\begin{equation}
\label{equca}
    CA^k(\boldsymbol{T}, \boldsymbol{I})=softmax(\frac{ (W_T^k \boldsymbol{T})(W_{I_1}^k \boldsymbol{I})^\top}{\sqrt{d}})(W_{I_2}^k\boldsymbol{I})^\top
\end{equation}
where $W_T^k $, $W_{I_1}^k$ and $W_{I_2}^k \in \mathbb{R}^{d \times d}$ are attention parameters in \textit{k-th} cross-attention layer. 

\subsection{FashionSAP Pre-training Tasks}
\label{subsec:pretraintask}


\subsubsection{Fashion Symbol Image Similarity (FSIS)}
This task makes the model capture the features from both text and image by maximizing the similarity between the image and the fashion symbol. In this task, the fashion symbol is concatenated between \verb'[CLS]' and the description tokens as shown in the upper-left of \cref{fig:modeloverview}. 
Let $\boldsymbol{v}_{symbol}$ denote the feature vector of the fashion symbol in the text side and $\boldsymbol{v}_{img}$ denote the feature vector of the image side. We use an adaptive layer $Adp(\cdot)$ to project the feature vector into adapted latent space. Let $\hat{\boldsymbol{v}}_{symbol} = \operatorname{norm}(Adp(\boldsymbol{v}_{symbol}))\in \mathbb{R}^{d_1}$ denote the adapted fashion symbol feature and  $\hat{\boldsymbol{v}}_{img} = \operatorname{norm}(Adp(\boldsymbol{v}_{img})\in \mathbb{R}^{d_1}$ denote the adapted image feature and $d_1$ is the dimension of adapted latent space.

The similarity between the fashion symbols and images is measured by modified vector cosine distance as \cref{equsis}
\begin{equation}
\label{equsis}
    \mathcal{L}_{fsis}=\frac{1}{B}{[ 1 - \sum_{b=1}^B {\frac{1}{2}{[\hat{\boldsymbol{v}}_{img}^b (\hat{\boldsymbol{v}}_{symbol}^b)^\top + 1]}}]}
\end{equation}
where $\operatorname{norm}$ is the normalize function, $B$ is the size of mini-batch, $\hat{\boldsymbol{v}}_{img}^b$ is the \textit{b-th} image adapted feature vector and $\hat{\boldsymbol{v}}_{symbol}^b$ is the \textit{b-th}  fashion symbol feature vector.

\subsubsection{Prompt Token Prediction (PTP)}
The goal of the PTP task is to improve the capacity of the model for learning from fine-grained attributes through predicting the correct token under a prompt. 
In this task, we choose a proper template $\mathcal{T}$ 
and use blank tokens to randomly hold the places of the name or value tokens with a probability of 0.5 to generate attribute input. This task minimizes the cross-entropy loss($G$). In addition, we use masked language modeling (MLM) task in model pre-training with loss calculated by $G$ as well.
So we merge these two losses as \cref{equptp}
\begin{equation}
\label{equptp}
    \mathcal{L}_{ptp} = \mathbb{E}_{(\boldsymbol{T}_{ptp}, \boldsymbol{I})\thicksim D}G(\boldsymbol{y}_{ptp}, \boldsymbol{g}_{ptp}(\boldsymbol{H}_{ptp}))
\end{equation}
where $\boldsymbol{H}_{ptp} = [\boldsymbol{H}_{mlm}\oplus \boldsymbol{H}_{pmt}]$ and $\oplus$ means the concatenation between two sequences, $\boldsymbol{y}_{ptp} = [\boldsymbol{y}_{mlm}\oplus \boldsymbol{y}_{pmt}]$ and $\boldsymbol{H}_{mlm},\boldsymbol{H}_{pmt}$ are hybrid features generated by feature fusion module with masked tokens and prompt tokens input respectively. $\boldsymbol{y}_{ptp}$ is the ground-truth and $\boldsymbol{g}_{ptp}(\boldsymbol{H}_{ptp})$ is the predicted probability distribution of the prompt token prediction task.

\subsubsection{Token Replace Prediction (TRP)}
In this task, first, we choose some tokens (ratio of 0.15) from the caption and one of the attribute values. Then, half of the chosen tokens are replaced by the antonyms searched by WordNet\cite{wrodnet} like \cite{replace} and the other half are replaced by random tokens from the vocabulary. This task aims at predicting whether the input tokens are substituted (labels 0 or 1).
The loss is shown in \cref{equtrp}
\begin{equation}
\label{equtrp}
\mathcal{L}_{trp} = \mathbb{E}_{(\boldsymbol{T}_{trp}, I)\thicksim D}G(\boldsymbol{y}_{trp}, \boldsymbol{g}_{trp}(\boldsymbol{H}_{trp}))
\end{equation}
$\boldsymbol{y}_{trp}$ is the ground-truth binary label and $\boldsymbol{g}_{trp}(\boldsymbol{H}_{trp})$ is the predicted probability distribution of the replacement task.

\subsubsection{Image Text Similarity (ITS)} This task aims at measuring the similarity between the text and the image. 
We use momentum contrastive learning \cite{contrastivelearning,MoCo,ALBEF} in this task to take full advantage of text-image pairs. As momentum contrastive learning requires mirror encoders for momentum updating, the vector $\boldsymbol{v}_{cls}$ denotes the whole semantics from the text and $\boldsymbol{v}_{cls}^{\prime}$ is the corresponding vector generated by momentum text encoder. Let vector $\boldsymbol{v}_{img}$ denote the whole feature from the image and $\boldsymbol{v}_{i}^{\prime}$ is generated by the momentum image encoder. The momentum distillation \cite{ALBEF,BLIP} is also used for label smoothing. For each pair of text and image, the similarities between them are \cref{equsimti} and \cref{equsimit}
\begin{equation}
\label{equsimti}
    \operatorname{sim}(\boldsymbol{T},\boldsymbol{I})=\operatorname{norm}(W_T\boldsymbol{v}_{cls})\operatorname{norm}(W_I \boldsymbol{v}_{img}^{\prime})^\top
\end{equation}
\begin{equation}
\label{equsimit}
     \operatorname{sim} (\boldsymbol{I},\boldsymbol{T})=\operatorname{norm}(W_I\boldsymbol{v}_{img})\operatorname{norm}(W_T \boldsymbol{v}_{cls}^{\prime})^\top
\end{equation}
where the $W_T$ and $W_I \in \mathbb{R}^{(d \times d)}$ are transfer weights to unify feature representations. The similarity between images and texts is measured by $\boldsymbol{g}_{i2t}$ and $\boldsymbol{g}_{t2i}$ and for \textit{k-th} image and text as \cref{equggi2t} ans \cref{equggt2i}
\begin{equation}
\label{equggi2t}
    g_{i2t}^k(\boldsymbol{I})=\frac{\exp(\operatorname{sim}(\boldsymbol{I}, \boldsymbol{T}^{k})/ \tau)}{\sum_{m=1}^M \exp(\operatorname{sim}(\boldsymbol{I},\boldsymbol{T}^m))}
\end{equation}
\begin{equation}
\label{equggt2i}
g_{t2i}^k(\boldsymbol{T})=\frac{\exp(\operatorname{sim}(\boldsymbol{T}, \boldsymbol{I}^{k})/ \tau)}{\sum_{m=1}^M \exp(\operatorname{sim}(\boldsymbol{T},\boldsymbol{I}^m))}
\end{equation}
where $\tau$ is a temperature parameter. The loss of the similarity of image and text is as \cref{equits}
\begin{equation}
\label{equits}
  \begin{split}
  \mathcal{L}_{its}=\frac{1}{2}\mathbb{E}_{(\boldsymbol{T},\boldsymbol{I})\sim D}[&G(\boldsymbol{y}_{i2t}(\boldsymbol{I}), \boldsymbol{g}_{i2t}(\boldsymbol{I})) + \\
  &G(\boldsymbol{y}_{t2i}(\boldsymbol{T}), \boldsymbol{g}_{t2i}(\boldsymbol{T}))]
  \end{split}
\end{equation}
where $\boldsymbol{y}_{i2t}(\boldsymbol{I})$ and $\boldsymbol{y}_{t2i}(\boldsymbol{T})$ denote the label of the similarity between images and texts.

\begin{figure}
  \centering
  \includegraphics[width=\linewidth]{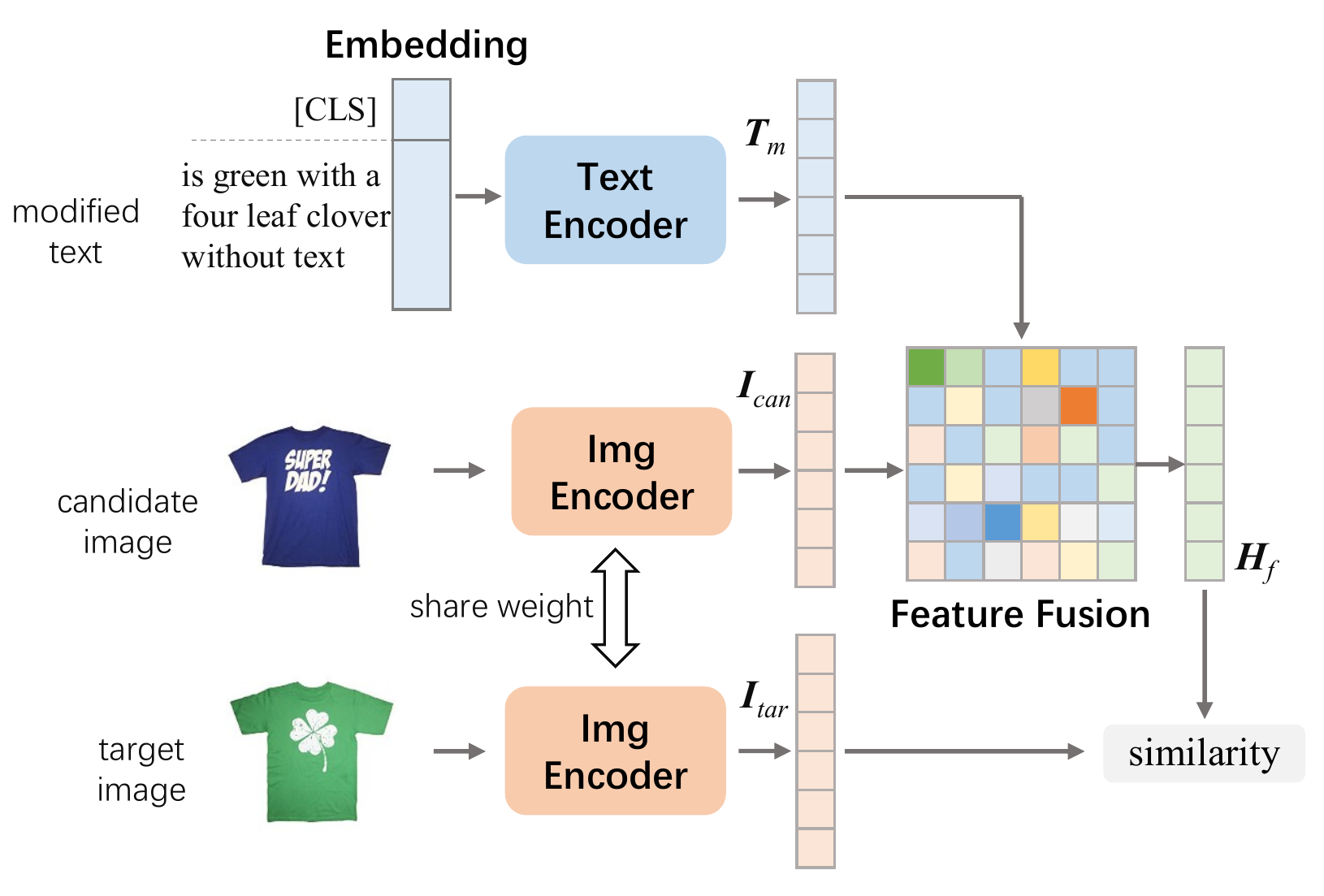}
  \caption{Model structure for TMIR task.}
  \label{fig:tmirmodel}
\end{figure}

\begin{table*}
  \centering
  \setlength\tabcolsep{12pt}
  \begin{tabular}{crclrclc}
    \toprule
    \multirow{2}{*}{Methods} & \multicolumn{3}{c}{I2T} & \multicolumn{3}{c}{T2I} & \multirow{2}{*}{Mean}\\
    \cmidrule(r){2-4} \cmidrule(r){5-7}
    ~ & R@1 & R@5 & R@10 & R@1 & R@5 & R@10 & R@1 \\
    \midrule
    VL-BERT\cite{vlbert} & 19.26 & 39.90 & 46.05 & 22.63 & 36.48 & 48.52  & 20.95\\
    ViLBERT\cite{lu2019vilbert} & 20.97 & 40.49 & 48.21 & 21.12 & 37.23 & 50.11  & 21.05\\
    Image-BERT\cite{qi2020imagebert} & 22.76 & 41.89 & 50.77 & 24.78 & 45.20 & 55.90  & 23.77\\
    OSCAR\cite{li2020oscar} & 23.39 & 44.67 & 52.55 & 25.10 & 49.14 & 56.68  & 24.25\\
    FashionBERT\cite{fashionbert} & 23.96 & 46.31 & 52.12 & 26.75 & 46.48 & 55.74  & 25.36\\
    KaleidoBERT\cite{zhuge2021kaleidobert} & 27.99 & 60.09 & 68.37 & 33.88 & 60.60 & 68.59  & 30.94\\
    EI-CLIP\cite{EICLIP} & 38.70 & 72.20 & 84.25 & 40.06 & 71.99 & 82.90 & 39.38\\ 
    CommerceMM\cite{yu2022commercemm} & 41.60 & 64.00 & 72.80 & 39.60& 61.50 & 72.70& 62.75 \\ 
    ALBEF\cite{ALBEF} & 63.97 & 88.92 & 94.41 & 60.52 & 84.99 & 91.45  & 62.20\\
    FashionViL\cite{han2022fashionvil} & 65.54 & 91.34 & 96.30 & 61.88 & 87.32 & 93.22  & 63.71\\
    \midrule
    FashionSAP(Resnet50) & 67.23 & 91.30 & 96.41 & 64.11 & 88.24 & 94.31 & 65.67 \\
    FashionSAP(ViT-B16) & 71.14 & 92.21 & 96.52 & 69.07 & 89.81 & 94.75 & 70.11 \\
    FashionSAP & \textbf{73.14} & \textbf{92.80} & \textbf{96.87} & \textbf{70.12} & \textbf{91.76} & \textbf{96.38}  & \textbf{71.63}\\
    \bottomrule
  \end{tabular}
  \caption{Cross-modal retrieval result on FashionGen\cite{fashiongen} in the sub set of evaluation following previous work. }
  \label{tab:retrievalsub}
\end{table*}

\begin{table*}
  \centering 
  \setlength\tabcolsep{12pt}
  \begin{tabular}{cccccccc}
    \toprule
    \multirow{2}{*}{Methods} & \multicolumn{3}{c}{I2T} & \multicolumn{3}{c}{T2I} & \multirow{2}{*}{Mean}\\
    \cmidrule(r){2-4} \cmidrule(r){5-7}
    ~ & R@1 & R@5 & R@10 & R@1 & R@5 & R@10 & R@1 \\
    \midrule
    EI-CLIP\cite{EICLIP}& 25.70 & 54.50 & 66.80 & 28.40 & 57.10 & 69.40 & 27.05 \\
    ALBEF\cite{ALBEF}& 41.68 & 67.39 & 75.50 & 50.95 & 75.36 & 84.15 & 46.32 \\
    FashionViL\cite{han2022fashionvil}& 42.88 & 71.57 & 80.55 & 51.34 & 75.42 & 84.57 & 47.11 \\
    \midrule
    FashionSAP(Resnet50) & 44.92 & 71.49 & 81.64 & 52.45 & 76.63 & 84.71 & 48.69 \\
    FashionSAP(ViT-B16) & 50.34 & 74.34 & 81.67 & 58.43 & 80.06 & 87.02 & 54.39 \\
    FashionSAP & \textbf{54.43} & \textbf{77.30} & \textbf{83.15} & \textbf{62.82} & \textbf{83.96} & \textbf{90.16}  & \textbf{58.63}\\
    \bottomrule
  \end{tabular}
  \caption{Cross-modal retrieval result on FashionGen\cite{fashiongen} with full evaluation}
  \label{tab:retrievalfull}
\end{table*}

\subsubsection{Image Text Match (ITM)}
In the task of image text match, the first vector of hybrid feature $H_{0}$ is sent to match head to predict the probability of text-image pair. The loss of this task is \cref{equitm}
\begin{equation}
\label{equitm}
    \mathcal{L}_{itm} =  \mathbb{E}_{(T, I)\thicksim D}G(\boldsymbol{y}_{itm}, \boldsymbol{g}_{itm}(H_0))
\end{equation}
where $\boldsymbol{g}_{itm}$ denote the predicted probability distribution by match head and $\boldsymbol{y}_{itm}$ denote the label(1 or 0) of image and text matching. The label is positive if the text-image is matched and negative if mismatched.

The complete pre-training objective of FashionSAP is the combination of the motioned terms above as \cref{eq.total},
\begin{equation}
\label{eq.total}
\mathcal{L}=\mathcal{L}_{fsis} + \mathcal{L}_{ptp} + \mathcal{L}_{trp} + \mathcal{L}_{its} +\mathcal{L}_{itm}
\end{equation}
The model is optimized end-to-end on the pre-training datasets by minimizing $\mathcal{L}$.


\section{Experiments}
\label{sec:expriments}
\begin{table*}
  \centering
    \setlength\tabcolsep{10pt}
  \begin{tabular}{ccccccccc}
    \toprule
    \multirow{2}{*}{Methods} & \multicolumn{2}{c}{Dress} & \multicolumn{2}{c}{Toptee} & \multicolumn{2}{c}{Shirt} & \multicolumn{2}{c}{Mean}\\
    \cmidrule(r){2-3} \cmidrule(r){4-5} \cmidrule(r){6-7} \cmidrule(r){8-9}
    ~ & R@10 & R@50 & R@10 & R@50 & R@10 & R@50 & R@10 & R@50 \\
    \midrule
    CIRR\cite{cirr} &17.45 &40.41 &21.64 &45.38 &17.53 &38.81 &18.87 &41.53 \\
    VAL\cite{val}  & 22.53 & 44.00 & 27.53 & 51.68  & 22.38 & 44.15 & 24.15 &46.61 \\
    CosMo\cite{lee2021cosmo} & 25.64 & 50.30 & 29.21 & 57.46  & 24.90 & 49.18 &26.58 &52.31 \\
    DCNet\cite{2021dcnet} & 28.95 &56.7 &30.44 &58.29 &23.95 &47.3 &27.78 &54.10 \\
    FashionVLP\cite{goenka2022fashionvlp} & 32.42 & 60.29 & 38.51 & 68.79 & 31.89 & 58.44 & 34.27 & 62.51 \\
    FashionViL\cite{han2022fashionvil} & 33.47 & 59.94  & 34.98 & 60.79&  25.17 & 50.39 & 31.21 & 57.04\\
    \midrule
    FashionSAP & \textbf{33.71} & \textbf{60.43} & \textbf{41.91} & \textbf{70.93} & \textbf{33.17} & \textbf{61.33} & \textbf{36.26} & \textbf{64.23}\\
    \bottomrule
  \end{tabular}
  \caption{Text modified image retrieval performance in FashionIQ\cite{fashioniq}}
  \label{tab:tmir}
\end{table*}

\subsection{Datasets}  We use the  FashionGen \cite{fashiongen} and FashionIQ\cite{fashioniq} datasets for pre-training and downstream tasks. FashionGen\cite{fashiongen} includes 320k pairs of text-image and 40k unique fashion items, which are shown as multiple images from multiple views. The detailed description and enumeration attributes are attached to all fashion items. FashionIQ\cite{fashioniq} dataset includes 77k unique fashion items and 18k modified text for text modified image retrieval task. We use the train set of FashionGen\cite{fashiongen} as pre-training data containing about 260k pairs of text-image. We evaluate downstream tasks text-to-image retrieval, image-to-text retrieval, category recognition and subcategory recognition in FashionGen \cite{fashiongen} and text modified image retrieval task in FashionIQ\cite{fashioniq}.

\subsection{Downstream Tasks and Results}

\textbf{Cross-modal Retrieval} We retrain only two losses, $\mathcal{L}_{its}$ and $\mathcal{L}_{itm}$ shown in \cref{fig:modeloverview} (lower-right) in this task. Cross-modal retrieval includes two tasks. One task is Image-to-Text (I2T), aiming to retrieve a matched text given a query image. Another task is Text-to-Image (T2I), which aims to retrieve a target image given a query text. 
We evaluate the performance of the model only by calculating the similarity between text and image following previous works. 

FashionSAP gets the SOTA performance as the comparable results shown in \cref{tab:retrievalsub}. We report the average result of 5 randomly chosen retrieval test sets and each of them contains 1k queries by following previous works. 
For each query in test sets, only one candidate is matched (positive), while the other 100 candidates are mismatched (negative) and chosen from the same subcategory. For the T2I task, there are 101 candidate images for each query text, and only one image in candidates is matched.

In order to test the performance of our model thoroughly, we also evaluate our model in the full test set of FashionGen\cite{fashiongen} in \cref{tab:retrievalfull} following \cite{han2022fashionvil,EICLIP}. 
Our model also gets the SOTA performance. Moreover, the differences between the results of our model and others are significant. 

We take a fine-tuning stage to the general VLP model (ALBEF) and report the results in \cref{tab:retrievalsub} and \cref{tab:retrievalfull}. We also provide the results of training FashionSAP from scratch with different image encoders following previous works.

\textbf{Category/Subcategory Recognition (CR\&SCR)} In this task, we only use cross-entropy loss for classification \cref{fig:modeloverview} (upper-right). This downstream tries to recognize the category and the subcategory, given the text and image of the fashion item. We extract the first vector of the fusion feature $H_0$ and input it to a linear layer to predict the category and the subcategory as shown in upper-right in \cref{fig:modeloverview}. FashionSAP gets the SOTA performance in both accuracy (Acc) and Macro-F as shown in \cref{tab:catereg}.

\begin{table}
  \centering \small
  \begin{tabular}{ccccc}
    \toprule
    \multirow{2}{*}{Methods} & \multicolumn{2}{c}{CR} & \multicolumn{2}{c}{SCR}\\
    \cmidrule(r){2-3} \cmidrule(r){4-5}
    ~ & Acc & Macro-F & Acc & Macro-F \\
    \midrule
    F-BERT\cite{fashionbert} &91.25 & 70.50 & 85.27& 62.00\\
    K-BERT\cite{zhuge2021kaleidobert} & 95.07 & 71.40 & 88.07 & 63.60\\
    F-ViL\cite{han2022fashionvil} & 97.48 & 88.60 & 92.23 & 83.02  \\
    \midrule
    FashionSAP & \textbf{98.34} & \textbf{89.84} & \textbf{94.33} & \textbf{87.67} \\
    \bottomrule
  \end{tabular}
  \caption{CR and SCR results on FashionGen\cite{fashiongen}.}
  \label{tab:catereg}
\end{table}

\textbf{Text Modified Image Retrieval (TMIR)}  This task aims at retrieving a target image of the fashion item by referring to the semantics of the query containing the features from a pair of candidate text-image while the text modifies some elements in the candidate image. As the original pre-training model can not be applied to this task directly, we design a new model structure for this task, shown in \cref{fig:tmirmodel}.  
The modified text is encoded into $\boldsymbol{T}_{m}$ meanwhile candidate image and target image are encoded into $\boldsymbol{I}_{can}$ and $\boldsymbol{I}_{tar}$. Then $\boldsymbol{T}_{m}$ and $\boldsymbol{I}_{can}$ are blended into hybrid feature $\boldsymbol{H}_{f}$. The cosine similarity between $\boldsymbol{H}_{f}$ and $\boldsymbol{I}_{tar}$ is the score between query and target and our model optimizes the similarity between them. Our model gets the SOTA performance compared with previous models, shown in \cref{tab:tmir}.


\subsection{Ablation Study}

\begin{figure*}
  \centering
  \includegraphics[width=0.8\textwidth]{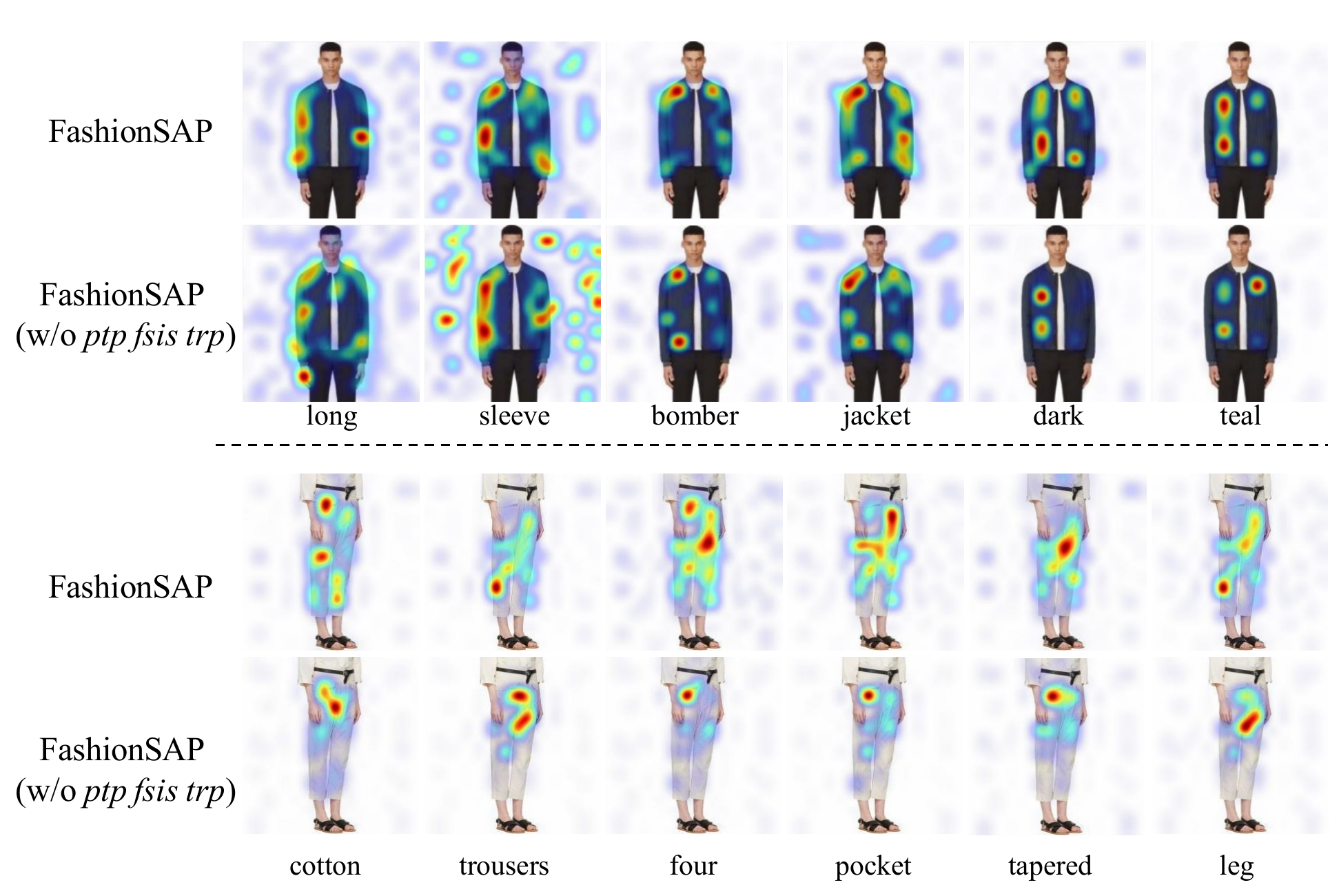}
  \caption{Instances of the comparison of Grad-CAM cross-attention maps for the 1st layer of the feature fusion module from FashionSAP (upper) and FashionSAP without three tasks (lower), Prompt Token Prediction task($ptp$), Fashion Symbol Image Similarity task($fsis$) and Token Replace Prediction task($trp$).}
  \label{fig:gradcam}
\end{figure*}

We evaluate the effectiveness of the proposed pre-training tasks in the section. For comparability, the settings in the same series of ablation are consistent. Considering the ITM task and ITS task are similar to general vision-language pre-training, we set the two tasks as basic ones and evaluate the three tasks proposed by this paper
in downstream tasks \cref{tab:Ablation}. For conciseness, we list only the index R@1 for both image-to-text and text-to-image tasks, index Macro-F for category (subcategory) recognition and index mean R@10 of three sets in FashionIQ\cite{fashioniq} for text modified image retrieval (TMIR).

\begin{table}
  \centering \small
  \setlength\tabcolsep{3.5pt}
  \begin{tabular}{ccccccccc}
    \toprule
    \multirow{2}{*}{$ptp$} & \multirow{2}{*}{$trp$}  & \multirow{2}{*}{$fsis$}& I2T & T2I & CR & SCR & TMIR\\
    ~ & ~& ~& R@1 &R@1 &Macro-F &Macro-F &R@10\\
    \midrule
    ~ & ~ & ~ & 43.84 & 53.24 & 84.50 & 84.42 & 30.02 \\ 
    \checkmark & ~ & ~ & 51.99 & 53.78 & 86.32 & 86.03 & 34.40 \\
    \checkmark & \checkmark & ~ &52.09 & 55.54 & 86.51 & 86.65 & 35.01\\ 
     \checkmark & \checkmark & \checkmark & \textbf{54.43} & \textbf{62.82} & \textbf{89.84} & \textbf{87.67} & \textbf{36.26} \\
    \bottomrule
  \end{tabular}
  \caption{Ablation study results for proposed tasks($ptp$, $fsis$, $trp$) on five downstream tasks.}
  \label{tab:Ablation}
\end{table}

As we can see from the results of the ablation study in \cref{tab:Ablation}, the loss $fsis$ brings a distinct improvement for T2I task as the fashion symbol is an essential structure capturing implicit semantics from the text side to the image side. The loss $ptp$ brings a distinct improvement for I2T task because the prompted fine-grained attributes are encoded as text tokens and share the same embedding layer with text. The loss $trp$ also brings an improvement in downstream tasks as the model learns synonym characteristics through this task.
\subsection{Fine-grained Alignment Analysis}

We choose two instances from FashionGen\cite{fashiongen} and show the cross-attention map in the T2I task using the Grad-CAM method \cref{fig:gradcam} to visualize the improvement of attention score. For each instance, we list the Grad-CAM visualizations from FashionSAP and FashionSAP without losses $ptp, fsis, trp$. Compared with the instances without proposed methods, FashionSAP concentrates on the corresponding region precisely. 
The two instances show that FashionSAP pays proper attention to the whole region of the object(\eg \verb|trousers|, \verb|leg|) rather than the sub-region. FashionSAP can also find all positions of \verb|pockets| in the attention maps rather than only one.

\subsection{Implementation Details}
The text encoder is the front 6-layer transformer of BERT-base\cite{BERT}, the image encoder is ViT-B16\cite{vit}. The feature fusion module is a 6-layer transformer. The feed-forward neural network implements the adapters, both on the text and image side. 
The FashionSAP is initialized by the checkpoint from ALBEF\cite{ALBEF} except for the results trained from scratch. 
The prompt predictor is a multi-layer feed-forward neural network. An AdamW\cite{adamw} optimizer is adopted with a learning rate $6e-5$. The batch size is 16 with momentum queue size 65535. The size of input images is $256 \times 256$. For training costs, we perform the pre-training stage in 8 Tesla V100*32G GPUs for 20 hours and fine-tuning stage for 10 hours. We randomly choose the attribute name or attribute value and replace them with their synonyms searched by WordNet\cite{wrodnet} for raw data preprocessing.

%

\section{Conclusion}
\label{sec:conclusion}
This paper introduced a fine-grained fashion VLP model for based on fashion symbols and attributes prompt. We used nine fashion symbols and attributes prompt to enhance the model to capture multi-modal fine-grained semantics. The comparative results and ablation study demonstrated that the FashionSAP was effective in learning fashion representation and outperforms SOTA models significantly.

Several future directions  could be considered. Our main goal was to show the potential of the attribute prompt framework to learn fine-grained fashion representation. The fashion symbol only considered category attributes and diversified symbols could be proposed. 

\section{Acknowledegements}
We thank the reviewers for their thoughtful and constructive comments. This work was supported in part by the National Key R\&D Program of China(2022ZD0116002), Natural Science Foundation of China (62276075, 61872113), Science and Technology Planning Project of Shenzhen (JCYJ20190806112210067), Huawei Technologies Co., Ltd. and Guangdong Provincial Key Laboratory of Novel Security Intelligence Technologies, China (2022B1212010005).
 %

{\small
\bibliographystyle{ieee_fullname}
\bibliography{egbib}
}

\end{document}